\documentclass[11pt,letterpaper]{article}
\usepackage{emnlp2017}
\usepackage{times}
\usepackage{latexsym}
\usepackage{amsmath}
\usepackage{graphicx}
\usepackage{booktabs}
\usepackage{subfig}
\usepackage{url}
\usepackage{tabularx}
\usepackage{multirow}
\usepackage{verbatim}
\usepackage{hyperref}

\emnlpfinalcopy

\title{Neural Machine Translation with Word Predictions}

\author{Rongxiang Weng, Shujian Huang, Zaixiang Zheng, Xinyu Dai \and Jiajun Chen \\
         State Key Laboratory for Novel Software Technology\\
         Nanjing University\\
         Nanjing 210023, China \\
         {\tt \{wengrx, huangsj, zhengzx, daixy, chenjj\}@nlp.nju.edu.cn}}
         
\date{}

\begin{document}

\maketitle

\begin{abstract}
In the encoder-decoder architecture for neural machine translation (NMT), the hidden states of the recurrent structures in the encoder and decoder carry the crucial information about the sentence.
These vectors are generated by parameters which are updated by back-propagation of translation errors through time.
We argue that propagating errors through the end-to-end recurrent structures are not a direct way of control the hidden vectors. 
In this paper, we propose to use word predictions as a mechanism for direct supervision. More specifically, we require these vectors to be able to predict the vocabulary in target sentence. Our simple mechanism ensures better representations in the encoder and decoder without using any extra data or annotation. It is also helpful in reducing the target side vocabulary and improving the decoding efficiency. Experiments on Chinese-English and German-English machine translation tasks show BLEU improvements by 4.53 and 1.3, respectively.
\end{abstract}

\section{Introduction}
The encoder-decoder based neural machine translation (NMT) models~\cite{sutskever2014sequence,Cho2014Learning} have been developing rapidly. 
\newcite{sutskever2014sequence} propose to encode the source sentence as a fixed-length vector representation, based on which the decoder generates the target sequence, where both the encoder and decoder are recurrent neural networks~(RNN)~\cite{sutskever2014sequence} or their variants~\cite{Cho2014Learning,Chung2014Empirical,Bahdanau2015Neural}. 
In this framework, the fixed-length vector plays the crucial role of transitioning the information of the sentence from the source side to the target side.

Later, attention mechanisms are proposed to enhance the source side representations~\cite{Bahdanau2015Neural,Luong2015Effective}. 
The source side context is computed at each time-step of decoding, based on the attention weights between the source side representations and the current hidden state of the decoder. However, the hidden states in the recurrent decoder still originate from the single fixed-length representation~\cite{Luong2015Effective}, or the average of the bi-directional representations~\cite{Bahdanau2015Neural}. Here we refer to the representation as \emph{initial state}. 

Interestingly, \newcite{Britz2017Massive} find that the value of initial state does not affect the translation performance, and prefer to set the initial state to be a zero vector. On the contrary, we argue that initial state still plays an important role of translation, which is currently neglected. We notice that beside the end-to-end error back propagation for the initial and transition parameters, there is no direct control of the initial state in the current NMT architectures. Due to the large number of parameters, it may be difficult for the NMT system to learn the proper sentence representation as the initial state. Thus, the model is very likely to get stuck in local minimums, making the translation process arbitrary and unstable. 

In this paper, we propose to augment the current NMT architecture with a word prediction mechanism. More specifically, we require the initial state of the decoder to be able to predict all the words in the target sentence. In this way, there is a specific objective for learning the initial state. Thus the learnt source side representation will be better constrained. We further extend this idea by applying the word predictions mechanism to all the hidden states of the decoder. So the transition between different decoder states could be controlled as well. 

Our mechanism is simple and requires no additional data or annotation. The proposed word predictions mechanism could be used as a training method and brings no extra computing cost during decoding. 

Experiments on the Chinese-English and German-English translation tasks show that both the constraining of the initial state and the decoder hidden states bring significant improvement over the baseline systems. Furthermore, using the word prediction mechanism on the initial state as a word predictor to reduce the target side vocabulary could greatly improve the decoding efficiency, without a significant loss on the translation quality. 

\section{Related Work}
\label{relating works}
Many previous works have noticed the problem of training an NMT system with lots of parameters. Some of them prefer to use the dropout technique~\cite{Srivastava2014Dropout,Luong2015Effective,Fandong2016Interactive}. Another possible choice is to ensemble several models with random starting points~\cite{sutskever2014sequence,Jean2015On,Luong2016Achieving}. Both techniques could bring more stable and better results. But they are general training techniques of neural networks, which are not specifically targeting the modeling of the translation process like ours. We will make empirical comparison with them in the experiments.

The way we add the word prediction is similar to the research of multi-task learning. \newcite{Dong2015Multi} propose to share an encoder between different translation tasks. \newcite{Luong2015Multi} propose to jointly learn the translation task for different languages, the parsing task and the image captioning task, with a shared encoder or decoder. \newcite{Zhang2016Exploiting} propose to use multitask learning for incorporating source side monolingual data. Different from these attempts, our method focuses solely on the current translation task, and does not require any extra data or annotation.

In the other sequence to sequence tasks, \newcite{SuzukiN17RNN} propose the idea for predicting words by using encoder information. However, the purpose and the way of our mechanism are different from them.

The word prediction technique has been applied in the research of both statistical machine translation~(SMT)~\cite{Bsngalore2007Statistical,Mauser2009Extending,jeong2010discriminative,monz2014word} and NMT~\cite{Haitao2016vocabulary,LHostisGA2016Vocabulary}. In these research, word prediction mechanisms are employed to decide the selection of words or constrain the target vocabulary, while in this paper, we use word prediction as a control mechanism for neural model training.

\begin{figure*}[ht]
\centering
\includegraphics[scale = 0.5]{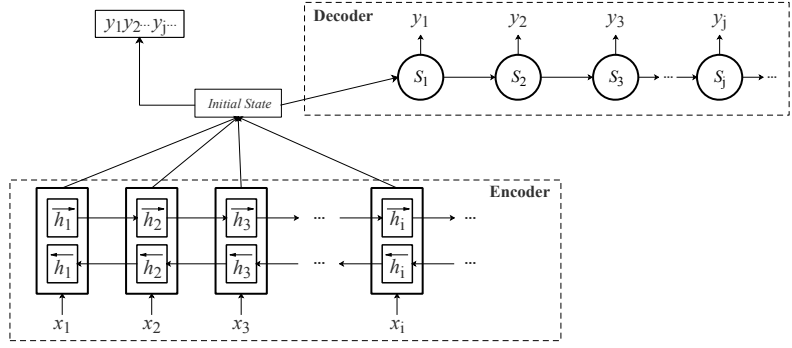}
\caption{\label{predict one} The NMT model with word prediction for the initial state.} 
\end{figure*}

\section{Notations and Backgrounds}
\label{framework}
We present a popular NMT framework with the encoder-decoder architecture~\cite{Cho2014Learning,Bahdanau2015Neural} and the attention networks~\cite{Luong2015Effective}, based on which we propose our word prediction mechanism.

Denote a source-target sentence pair as $ \{\textbf{x}, \textbf{y}\} $ from the training set, where $\textbf{x}$ is the source word sequence $(x_{1},x_{2}, \cdots,x_{|\textbf{x}|})$ and $\textbf{y}$ is the target word sequence $(y_{1},y_{2}, \cdots,y_{|\textbf{y}|})$, $|\textbf{x}|$ and $|\textbf{y}|$ are the length of $\textbf{x}$ and $\textbf{y}$, respectively.

In the encoding stage, a bi-directional recurrent neural network is used~\cite{Bahdanau2015Neural} to encode $ \textbf{x} $ into a sequence of vectors $ (\textbf{h}_{1},\textbf{h}_{2},\cdots,\textbf{h}_{|\textbf{x}|}) $. For each $x_i$, the representation $\textbf{h}_i$ is:
\begin{equation}
\textbf{h}_{i}=[\overrightarrow{\textbf{h}_{i}};\overleftarrow{\textbf{h}_{i}}]
\end{equation}
where $[\cdot;\cdot]$ denotes the concatenation of column vectors; $\overrightarrow{\textbf{h}_{i}}$ and $\overleftarrow{\textbf{h}_{i}}$ denote the hidden vectors for the word $x_i$ in the forward and backward RNNs, respectively.

The gated recurrent unit~(GRU) is used as the recurrent unit in each RNN, which is shown to have promising results in speech recognition and machine translation~\cite{Cho2014Learning}. 
Formally, the hidden state $\textbf{h}_{i}$ at time step $i$ of the forward RNN encoder is defined by the GRU function $g_{\overrightarrow{e}}(\cdot,\cdot)$, as follows:
\begin{align}
\overrightarrow{\textbf{h}}_{i}&=g_{\overrightarrow{e}}(\overrightarrow{\textbf{h}}_{i-1},\textbf{emb}_{x_{i}}) \label{eq-GRU} \\
&=(\textbf{1}-\overrightarrow{\textbf{z}}_{i})\odot \overrightarrow{\textbf{h}}_{i-1}+
\overrightarrow{\textbf{z}}_{i}\odot \overrightarrow{\textbf{h}'}_{i} \nonumber\\
\overrightarrow{\textbf{z}}_{i}&=\sigma(\overrightarrow{\textbf{W}}_{z}[\textbf{emb}_{x_{i}};
\overrightarrow{\textbf{h}}_{i-1}]) \\
\overrightarrow{\textbf{h}'}_{i}&=\text{tanh}(\overrightarrow{\textbf{W}}[\textbf{emb}_{x_{i}};
(\overrightarrow{\textbf{r}}_{i} \odot \overrightarrow{\textbf{h}}_{i-1})]) \\
\overrightarrow{\textbf{r}}_{i}&=\sigma (\overrightarrow{\textbf{W}_{r}}[\textbf{emb}_{x_{i}}; \overrightarrow{\textbf{h}}_{i-1}])
\end{align}

where  $ \odot $ denotes element-wise product between vectors and $ \textbf{emb}_{x_{i}}$ is the word embedding of the $x_{i}$. 
$\text{tanh}(\cdot)$ and $\sigma (\cdot)$ are the tanh and sigmoid transformation functions that can be applied element-wise on vectors, respectively. For simplicity, we omit the bias term in each network layer. The backward RNN encoder is defined likewise.

In the decoding stage, the decoder starts with the initial state $\textbf{s}_{0}$, which is the average of source representations~\cite{Bahdanau2015Neural}.

\begin{align}
\textbf{s}_{0} &= \sigma(\textbf{W}_{s}\frac{1}{|\textbf{x}|} \sum_{i=1}^{|\textbf{x}|}\textbf{h}_{i}) \label{eq-s0}
\end{align}

At each time step $j$, the decoder maximizes the conditional probability of generating the $ j $th target word, which is defined as:
\begin{align}
 P(y_{j}|y_{<j},\textbf{x})&=f_{d}(t_{d}([\textbf{emb}_{y_{j-1}};\textbf{s}_{j};\textbf{c}_{j}])) \label{eq-pt}\\
 f_{d}(\textbf{u})&=\text{softmax}(\textbf{W}_{f}\textbf{u}) \label{eq-f}\\ 
 t_{d}(\textbf{v})&=\text{tanh}(\textbf{W}_{t}\textbf{v}) \label{eq-t}
\end{align}
where $ \textbf{s}_{j} $ is the decoder's hidden state, which is computed by another GRU (as in Equation~\ref{eq-GRU}):
\begin{align}
\textbf{s}_{j} &= g_{d}(\textbf{s}_{j-1},[\textbf{emb}_{y_{j-1}};\textbf{c}_{j}])
\end{align}

and the context vector $ \textbf{c}_{j} $ is from the attention mechanism~\cite{Luong2015Effective}:
\begin{align}
\textbf{c}_{j}&=\sum_{i=1}^{|\textbf{x}|}a_{ji}\textbf{h}_{i}\\
a_{ji}&=\frac{\text{exp}(e_{ji})}{\sum_{k=1}^{|\textbf{x}|}\text{exp}(e_{jk})}\\
e_{ji}&=\text{tanh}(\textbf{W}_{att_{d}}[\textbf{s}_{j-1};\textbf{h}_{i}]).
\end{align}

\begin{figure*}[ht]
\centering
\includegraphics[scale = 0.5]{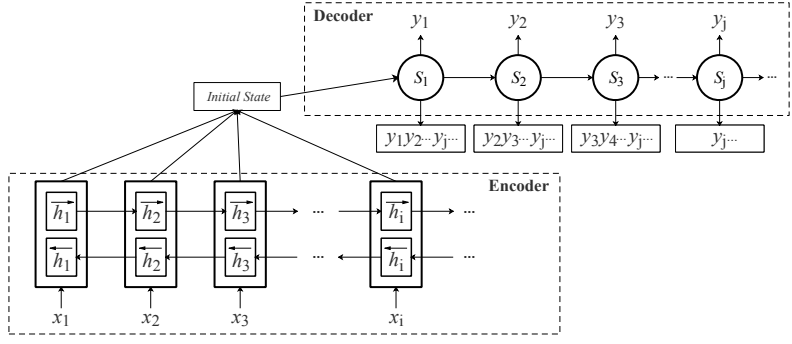}
\caption{\label{predict iteration} The NMT model with word predictions for the decoder's hidden states.} 
\end{figure*}

\section{NMT with Word Predictions}
\subsection{Word Prediction for the Initial State}
The decoder starts the generation of target sentence from the initial state $\textbf{s}_0$ (Equation~\ref{eq-s0}) generated by the encoder. Currently, the update for the encoder only happens when a translation error occurs in the decoder. The error is propagated through multiple time steps in the recurrent structure until it reaches the encoder. As there are hundreds of millions of parameters in the NMT system, it is hard for the model to learn the exact representation of source sentences. As a result, the values of initial state may not be exact during the translation process, leading to poor translation performances. 

We propose word prediction as a mechanism to control the values of initial state. The intuition is that since the initial state is responsible for the translation of whole target sentence, it should at least contain information of each word in the target sentence. Thus, we optimize the initial state by making prediction for all target words. For simplicity, we assume each target word is independent of each other.

Here the word prediction mechanism is a simpler sub-task of translation, where the order of words is not considered. The prediction task could be trained jointly with the translation task in a multi-task learning way~\cite{Luong2015Multi,Dong2015Multi,Zhang2016Exploiting}, where both tasks share the same encoder. In other words, word prediction for the initial state could be interpreted as an improvement for the encoder. We denote this mechanism as $\text{WP}_{\text{E}}$ .

As shown in Figure~\ref{predict one}, a prediction network is added to the initial state. We define the conditional probability of $\text{WP}_{\text{E}}$ as follows:
\begin{align}
P_{\text{WP}_{\text{E}}}(\textbf{y}|\textbf{x})&=\prod_{j=1}^{|\textbf{y}|}P_{\text{WP}_{\text{E}}}(y_{j}|\textbf{x}) \label{eq-pwpe}\\
P_{\text{WP}_{\text{E}}}(y_{j}|\textbf{x})&= f_{p}(t_{p}([\textbf{s}_{0};\textbf{c}_{p}]))
\end{align}
where $f_{p}(\cdot)$ and $t_p(\cdot)$ are the softmax layer and non-linear layer as defined in Equation~\ref{eq-f}-\ref{eq-t}, with different parameters; $\textbf{c}_{p}$ is defined similar as the attention network, so the source side information could be enhanced.
\begin{align}
\textbf{c}_{p}&=\sum_{i=1}^{|\textbf{x}|}a_{i}\textbf{h}_{i} \label{eq-s0attention}\\
a_{i}&=\frac{\text{exp}(e_{i})}{\sum_{k=1}^{|\textbf{x}|}\text{exp}(e_{k})}\\
e_{i}&=\text{tanh}(\textbf{W}_{att_{p}}[\textbf{s}_{0},\textbf{h}_{i}]).
\end{align}

\subsection{Word Predictions for Decoder's Hidden States}
Similar intuition is also applied for the decoder. Because the hidden states of the decoder are responsible for the translation of target words, they should be able to predict the target words as well. The only difference is that we remove the already generated words from the prediction task. So each hidden state in the decoder is required to predict the target words which remain untranslated. 

For the first state $\textbf{s}_1$ of the decoder, the prediction task is similar with the task for the initial state. Since then, the prediction is no longer a separate training task, but integrated into each time step of the training process. We denote this mechanism as $\text{WP}_{\text{D}}$.

As shown in Figure~\ref{predict iteration}, for each time step $j$ in the decoder, the hidden state $\textbf{s}_j$ is used for the prediction of  $(y_{j}, y_{j+1}, \cdots, y_{|\textbf{y}|} )$.
The conditional probability of $\text{WP}_{\text{D}}$ is defined as:
\begin{align}
P_{\text{WP}_{\text{D}}}(y_{j},y_{j+1},  & \cdots ,   y_{|\textbf{y}|} | y_{<j},\textbf{x}) \label{eq-pwpd} \\
= & \prod_{k=j}^{|\textbf{y}|}    P_{\text{WP}_{\text{D}}}(y_{k}|y_{<j},\textbf{x}) \nonumber\\
P_{\text{WP}_{\text{D}}}(y_{k}|y_{<j}, \textbf{x}) = & f_{d} (p(t_{d}([\textbf{emb}_{y_{j-1}};\textbf{s}_{j};\textbf{c}_{j}])))
\end{align}
where $f_{d}(\cdot)$ and $t_d(\cdot)$ are the softmax layer and non-linear layer as defined in Equation~\ref{eq-f}-\ref{eq-t}; $ p(\cdot) $ is another non-linear transformation layer, which prepares the current state for the prediction:
\begin{align}
p(\textbf{u})=\text{tanh}(\textbf{W}_{p}\textbf{u}).
\end{align}

\subsection{Training}
NMT models optimize the networks by maximizing the likelihood of the target translation $\textbf{y}$ given source sentence $\textbf{x}$, denoted by $L_\text{T}$.
\begin{equation}
L_{\text{T}}=\frac{1}{|\textbf{y}|}\sum_{j=1}^{|\textbf{y}|}\log P(y_{j}|y_{<j},\textbf{x})
\end{equation}
where $P(y_{j}|y_{<j},\textbf{x})$ is defined in Equation~\ref{eq-pt}.

To optimize the word prediction mechanism, we propose to add extra likelihood functions $L_{\text{WP}_{\text{E}}}$ and $L_{\text{WP}_{\text{D}}}$ into the training procedure. 


For the $\text{WP}_{\text{E}}$, we directly optimize the likelihood of translation and word prediction:
 \begin{align}
L_1&=L_{\text{T}}+L_{\text{WP}_{\text{E}}}\\
L_{\text{WP}_{\text{E}}}&=\log P_{\text{WP}_{\text{E}}}
\end{align}
where $ P_{\text{WP}_{\text{E}}} $ is defined in Equation~\ref{eq-pwpe}.

For the $ \text{WP}_{\text{D}} $, we optimize the likelihood as:
\begin{align}
L_2&=L_{\text{T}}+L_{\text{WP}_{\text{D}}}\\
L_{\text{WP}_{\text{D}}}&=\sum_{j=1}^{|\textbf{y}|}\frac{1}{|\textbf{y}|-j+1}\log P_{\text{WP}_{\text{D}}}
\end{align}
where $ P_{\text{WP}_{\text{D}}} $ is defined in Equation~\ref{eq-pwpd}; the coefficient of the logarithm is used to calculate the average probability of each prediction.

The two mechanisms could also work together, so that both the encoder and the decoder could be improved:
\begin{align}
L_3&=L_{\text{T}}+L_{\text{WP}_{\text{E}}}+L_{\text{WP}_{\text{D}}}.
\end{align}

\subsection{Making Use of the Word Predictor}
The previously proposed word prediction mechanism could be used only as a extra training objective, which will not be computed during the translation. Thus the computational complexity of our models for translation stays exactly the same. 

 On the other hand, using a  smaller and specific vocabulary for each sentence or batch will improve translation efficiency.  If the vocabulary is accurate enough, there is also a chance to improve the translation quality~\cite{Jean2015On,Haitao2016vocabulary,LHostisGA2016Vocabulary}. Our word prediction mechanism WP$_\text{E}$ provides a natural solution for generating a possible set of target words at sentence level. The prediction could be made from the initial state $\textbf{s}_0$, without using  extra resources such as word dictionaries, extracted phrases or frequent word lists, as in~\newcite{Haitao2016vocabulary}. 

\section{Experiments}
\label{experimental setup}

\subsection{Data}

We perform experiments on the Chinese-English (CH-EN) and German-English (DE-EN) machine translation tasks. For the CH-EN, the training data consists of about 8 million sentence pairs~\footnote{includes LDC2002E18, LDC2003E07, LDC2003E14, LDC2004E12, LDC2004T08, LDC2005T06, LDC2005T10, LDC2006E26 and LDC2007T09}. 
We use NIST MT02 as our validation set, and the NIST MT03, MT04 and MT05 as our test sets. 
These sets have 878, 919, 1597 and 1082 source sentences, respectively, with 4 references for each sentence.
For the DE-EN, the experiments trained on the standard benchmark WMT14, and it has about 4.5 million sentence pairs. We use newstest 2013 (NST13) as validation set, and newstest 2014(NST14) as test set.
These sets have 3000 and 2737 source sentences, respectively, with 1 reference for each sentence. Sentences were encoded using byte-pair encoding (BPE)~\cite{Britz2017Massive}.

\subsection{Systems and Techniques}
We implement a baseline system with the bi-directional encoder~\cite{Bahdanau2015Neural} and the attention mechanism~\cite{Luong2015Effective} as described in Section~\ref{framework}, denoted as baseNMT. Then our proposed word prediction mechanism on initial state and hidden states of decoder are implemented on the baseNMT system, denoted as ${\text{WP}_{\text{E}}}$ and ${\text{WP}_{\text{D}}}$, respectively. We denote the system use both techniques as ${\text{WP}_{\text{ED}}}$. We implement systems with variable-sized vocabulary following~\cite{Haitao2016vocabulary}. For comparison, we also implement systems with dropout (with dropout rate 0.5 on the output layer) and ensemble (ensemble of 4 systems at the output layer) techniques.

\subsection{Implementation Details}
Both our CH-EN and DE-EN experiments are implemented on the open source toolkit dl4mt~\footnote{\href{https://github.com/nyu-dl/dl4mt-tutorial}{https://github.com/nyu-dl/dl4mt-tutorial}}, with most default parameter settings kept the same. We train the NMT systems with the sentences of length up to 50 words. The source and target vocabularies are limited to the most frequent 30K words for both Chinese and English, respectively, with the out-of-vocabulary words mapped to a special token UNK.

The dimension of word embedding is set to 512 and the size of the hidden layer is 1024.
The recurrent weight matrices are initialized as random orthogonal matrices, and all the bias vectors as zero. 
Other parameters are initialized by sampling from the Gaussian distribution $\mathcal{N}(0, 0.01)$.
  
We use the mini-batch stochastic gradient descent~(SGD) approach to update the parameters, with a batch size of 32. The learning rate is controlled by AdaDelta~\cite{Zeiler2012ADADELTA}.

For efficient training of our system, we adopt a simple pre-train strategy. Firstly, the baseNMT system is trained. The training results are used as the initial parameters for pre-training our proposed models with word predictions.

For decoding during test time, we simply decode until the end-of-sentence symbol $eos$ occurs, using a beam search with a beam width of 5.

\subsection{Translation Experiments}
To see the effect of word predictions in translation, we evaluate these systems in case-insensitive IBM-BLEU~\cite{Papineni2002bleu} on both CH-EN and DE-EN tasks.

\begin{table*}[ht]
\centering
\begin{tabular}{c|c|ccc|c|c}
\hline
Models& MT02(dev)&MT03 & MT04 & MT05&Test Ave.&IMP\\
\hline
baseNMT&34.04&34.92&36.08&33.88&34.96& $-$ \\
\hline
WP$_\text{E}$&39.36&37.17&39.11&36.20&37.49&\textbf{+2.53}\\
\hline
WP$_\text{D}$&40.28&38.45&40.99& 37.90&39.11 &\textbf{+4.15}\\
\hline
WP$_\text{ED}$&40.25&39.50&40.91&38.05&39.49&\textbf{+4.53}\\
\hline
\end{tabular}
\caption{\label{bleu result chen} Case-insensitive 4-gram BLEU scores of baseNMT, WP$_\text{E}$, WP$_\text{D}$, WP$_\text{ED}$ systems on the CH-EN experiments. 
(The ``IMP'' column presents the improvement of test average compared to the baseNMT. )}
\end{table*}

The detailed results are show in the Table~\ref{bleu result chen} and Table~\ref{bleu result deen}. Compared to the baseNMT system, all of our models achieve significant improvements. On the CH-EN experiments, simply adding word predictions to the initial state (WP$_\text{E}$) already brings considerable improvements. The average improvement on test set is 2.53 BLEU, showing that constraining the initial state does lead to a higher translation quality.  Adding word predictions to the hidden states in the decoder (WP$_\text{D}$) leads to further improvements against baseNMT (4.15 BLEU), because WP$_\text{D}$ adds constraints to the state transitions through different time steps in the decoder. Using both techniques improves the baseline by 4.53 BLEU. 
On the DE-EN experiments, the improvement of WP$_\text{E} $ model is 0.41 BLEU and WP$_\text{D}$ model is 0.86 BLEU on test set. When use both techniques, the WP$_\text{ED}$ improves on the test set is 1.3 BLEU.

\begin{table}[t]
\centering
\begin{tabular}{l|c|c|c}
\hline
Models&NST13(dev)&NST14&IMP\\
\hline
baseNMT&23.56&20.68&$-$\\
\hline
WP$_\text{E}$&24.44&21.09&\textbf{+0.41}\\
\hline
WP$_\text{D}$&25.31&21.54&\textbf{+0.86}\\
\hline
WP$_\text{ED}$&25.97&21.98&\textbf{+1.3}\\
\hline
\end{tabular}
\caption{\label{bleu result deen}Case-insensitive 4-gram BLEU scores of baseNMT, WP$_\text{E}$, WP$_\text{D}$, WP$_\text{ED}$ systems on the DE-EN experiments. }
\end{table}

\begin{table}[t]
\centering
\begin{tabular}{l|c|c}
\hline
Models&Test&IMP\\
\hline
baseNMT&34.86&$-$\\
\hline
WP$_\text{ED}$&39.49&+4.53\\
\hline
\hline
baseNMT-dropout&37.02&+2.06\\
\hline
WP$_\text{ED}$-dropout&39.25&+4.29\\
\hline
\hline
baseNMT-ensemble(4)&37.71&+2.75\\
\hline
WP$_\text{ED}$-ensemble(4)&40.75&+5.79\\
\hline
\end{tabular}
\caption{\label{compare other models chen}Average case-insensitive 4-gram BLEU scores on the CH-EN experiments for baseNMT and WP$_\text{ED}$ systems, with the dropout and ensemble techniques.}
\end{table}

\begin{table}[t]
\centering
\begin{tabular}{l|c|c}
\hline
Models&Test&IMP\\
\hline
baseNMT&20.68&$-$\\
\hline
WP$_\text{ED}$&21.98&+1.3\\
\hline
\hline
baseNMT-dropout&21.62&+0.94\\
\hline
WP$_\text{ED}$-dropout&21.71&+1.03\\
\hline
\hline
baseNMT-ensemble(4)&21.58&+0.9\\
\hline
WP$_\text{ED}$-ensemble(4)&22.47&+1.79\\
\hline
\end{tabular}
\caption{\label{compare other models deen}Case-insensitive 4-gram BLEU scores on the DE-EN experiments for baseNMT and WP$_\text{ED}$ systems, with the dropout and ensemble techniques.}
\end{table}

We compare our models with systems using dropout and ensemble techniques. The results show in Table~\ref{compare other models chen} and~\ref{compare other models deen}.   
On the CH-EN experiments, the dropout method successfully improves the baseNMT system by 2.06 BLEU. However, it does not work on our WP$_\text{ED}$ system.  The ensemble technique improves the baseNMT system by 2.75 BLEU. It still improves WP$_\text{ED}$ by 1.26 BLEU, but the improvement is smaller than on the baseNMT. 
On the DE-EN experiments, the phenomenon of experiments is similar to CH-EN experiments. The baseNMT system improves 0.94 through dropout method and 0.9 BLEU through ensemble method. The dropout technique also does not work on WP$_\text{ED}$ and the ensemble technique improves 1.79 BLEU.
These comparisons suggests that our system already learns better and stable values for the parameters, enjoying some of the benefits of general training techniques like dropout and ensemble. 
Compared to dropout and ensemble, our method WP$_\text{ED}$ achieves the highest improvement against the baseline system  on both CH-EN and DE-EN experiments. Along with ensemble method, the improvement could be up to 5.79 BLEU and 1.79 BLEU respectively.

\subsection{Word Prediction Experiments}

Since we include an explicit word prediction mechanism during the training of NMT systems, we also evaluate the prediction performance on the CH-EN experiments to see how the training is improved.

For each sentence in the test set, we use the initial state of the given model to make prediction about the possible words. We denote the set of top $n$ words as $T_n$, the set of words in all the references as $R$. 
We define the precision, recall of the word prediction as follows:
\begin{align}
\text{precision}&=\frac{|T_n\cap R|}{|T_n|}*100\%\\
\text{recall}&=\frac{|T_n\cap R|}{|R|}*100\% 
\end{align}

We compare the prediction performance of baseNMT and WP$_\text{E}$. WP$_\text{ED}$ has similar prediction results with WP$_\text{E}$, so we omit its results. As shown in Table~\ref{prf}, baseNMT system has a relatively lower prediction precision, for example, 45\% in top 10 prediction. With an explicit training, the WP$_\text{E}$ could achieve a much higher precision in all conditions. Specifically, the precision reaches 73\% in top 10. This indicates that the initial state in WP$_\text{E}$ contains more specific information about the prediction of the target words, which may be a step towards better semantic representation, and leads to better translation quality. 

\begin{table}[t]
\centering
\begin{tabular}{*{4}{c|}c}
\hline
\multirow{2}{*}{\centering  top-$n$}&\multicolumn{2}{c|}{baseNMT}&\multicolumn{2}{c}{WP$_\text{E}$}\\
\cline{2-5}
&Prec.&Recall&Prec.&Recall\\
\hline
top-10&45\%&17\%&73\%&30\%\\
\hline
top-20&33\%&21\%&63\%&43\%\\
\hline
top-50&21\%&30\%&41\%&55\%\\
\hline
top-100&14\%&39\%&28\%&68\%\\
\hline
top-1k&2\%&67\%&4\%&89\%\\
\hline
top-5k&0.7\%&84\%&0.9\%&95\%\\
\hline
top-10k&0.4\%&90\%&0.5\%&97\%\\
\hline
\end{tabular}
\caption{\label{prf}Comparison between baseNMT and WP$_\text{E}$ in precision and recall for the different prediction size on the CH-EN experiments. 
}
 \end{table}
 
\begin{figure}[ht]
\centering
\includegraphics[scale=0.4]{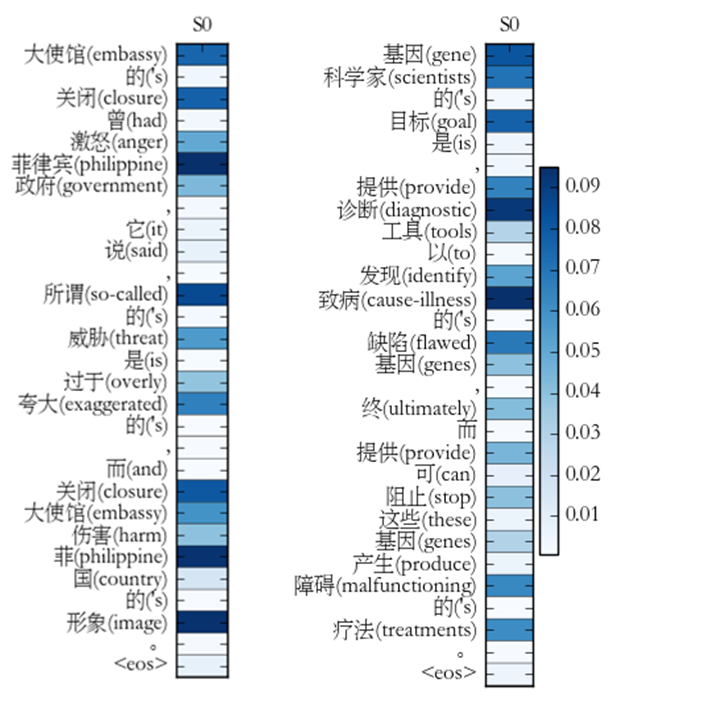}
\caption{Two examples of the attention heatmap between the initial state $\textbf{s}_0$ and the bi-directional representation of each source side word $\textbf{h}_i$ from the CH-EN test sets. (The English translation of each source word is annotated in the parentheses after it. )
}
\label{fg-s0attention}
\end{figure}

Because the total words in the references are limited (around 50), the precision goes down, as expected, when a larger prediction set is considered.
On the other hand, the recall of WP$_\text{E}$ is also much higher than baseNMT. When given 1k predictions,  WP$_\text{E}$ could successfully predict 89\% of the words in the reference. The recall goes up to 95\% with 5k predictions, which is only 1/6 of the current vocabulary.

\begin{figure}[ht]
\centering
\includegraphics[scale=0.34]{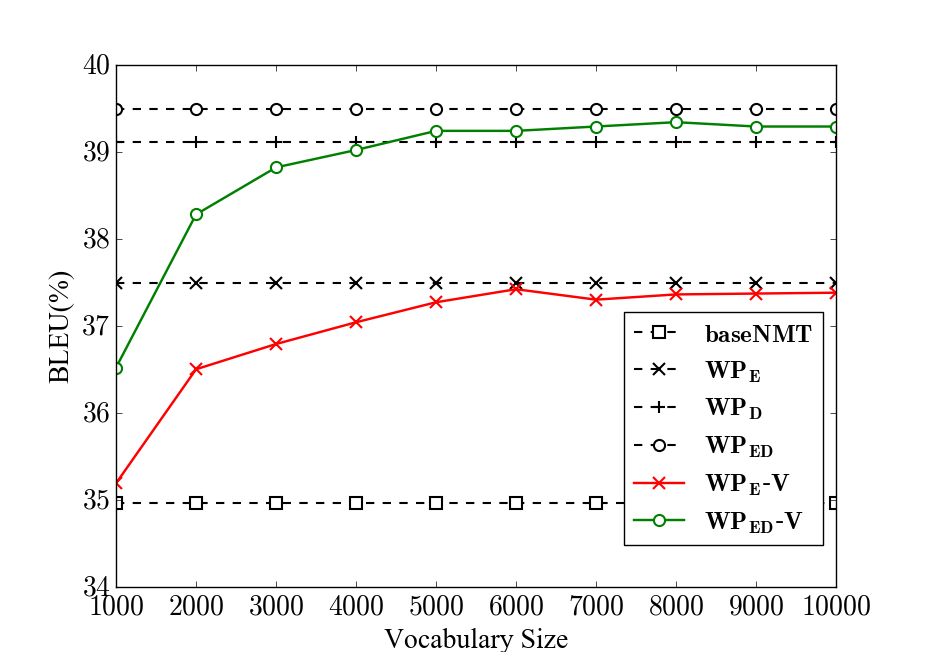}
\caption{BLEU scores with different vocabulary sizes for each sentence on the CH-EN experiments. (The performance of baseNMT, WP$_{\text{E}}$, WP$_{\text{D}}$, WP$_{\text{ED}}$ are plotted as horizontal lines for comparison.) 
}
\label{fg-vbleu}
\end{figure}

\begin{figure}[ht]
\centering
\includegraphics[scale=0.34]{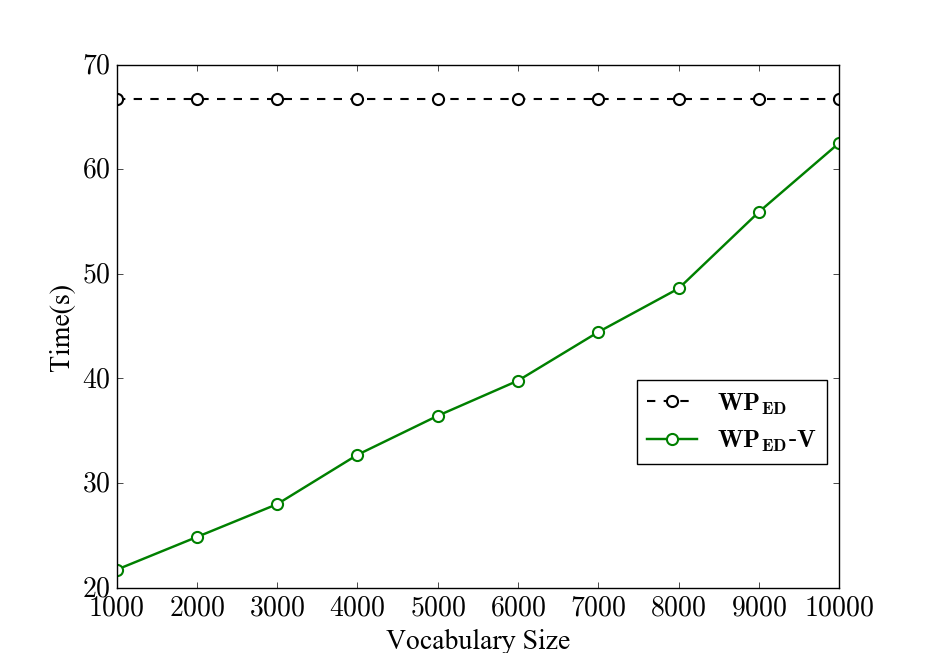}
\caption{Decoding time with different vocabulary sizes for each sentence on the CH-EN experiments. (The horizontal line shows the decoding time for the systems with fixed vocabulary. )
}
\label{fg-vtime}
\end{figure}

To analyze the process of word prediction, we draw the attention heatmap (Equation~\ref{eq-s0attention}) between the initial state $\textbf{s}_0$ and the bi-directional representation of each source side word $\textbf{h}_i$ for an example sentence. As shown in Figure~\ref{fg-s0attention}, both examples show that the initial states have a very strong attention with all the content words in the source sentence. The blank cells are mostly functions words or high frequent tokens such as ``de('s)", ``shi(is)", ``er(and)", ``ta(it)", comma and period.  This indicates that the initial state successfully encodes information about most of the content words in the source sentence, which contributes for a high prediction performance and leads to better translation.

\subsection{Improving Decoding Efficiency}

To make use of the word prediction, we conduct experiments using the predicted vocabulary, with different vocabulary size (1k to 10k) on the CH-EN experiments, denoted as WP$_\text{E}$-V and WP$_\text{ED}$-V. The comparison is made in both translation quality and decoding time. As all our models with fixed vocabulary size have exactly the same number of parameters for decoding (extra mechanism is used only for training), we only plot the decoding time of the WP$_\text{ED}$ for comparison. Figure~\ref{fg-vbleu} and \ref{fg-vtime} show the results.

When we start the experiments with top 1k vocabulary (1/30 of the baseline settings), the translation quality of both WP$_\text{E}$-V and WP$_\text{ED}$-V are already higher than the baseNMT; while their decoding time is less than 1/3 of an NMT system with 30k vocabulary. When the size of vocabulary increases, the translation quality improves as well. With a 6k predicted vocabulary (1/5 of the baseline settings), the decoding time is about 60\% of a full-vocabulary system; the performances of both systems with variable size vocabulary are comparable their corresponding fixed-vocabulary systems, which is higher than the baseNMT by 2.53 and 4.53 BLEU, respectively.

\begin{table*}[t!]
\centering
\begin{tabular}{ m{50pt} | m{380pt} }
\hline
source & \includegraphics[scale=1.05]{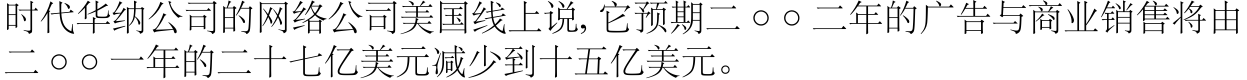} \\ 
\hline
reference & america online , the internet arm of time warner conglomerate , said it expects \underline{advertising} and commerce revenue to decline from us \$ 2.7 billion in 2001 to us \$ \underline{1.5} in 2002 .\\
\hline
baseNMT & in the \emph{us line} , \emph{the internet company 's internet company said on the internet} that it expected that the business sales in 2002 would fall from \$ UNK billion to \$ UNK billion in 2001 . \\
\hline
baseNMT +dropout & on \emph{the united states line} , \emph{UNK 's} internet company said \emph{on the internet} that it expects to reduce the annual \textbf{advertising and commercial} sales from \$ UNK billion in 2001 to \$ \textbf{1.5} billion .\\
\hline
baseNMT +ensemble & in the \emph{us line} , \emph{the internet company 's internet company} said that it expected that the \textbf{advertising and commercial} sales volume for 2002 would be reduced from us \$ UNK billion to us \$ \textbf{1.5} billion in 2001 .\\
\hline
WP$_\text{ED}$ & \textbf{the internet company} \emph{of} \textbf{time warner inc. }, the \emph{us} online , said that it expects that \textbf{the advertising and commercial} sales in 2002 will decrease from \$ UNK billion in 2001 to us \$ \textbf{1.5} billion .\\
\hline
\end{tabular}
\caption{\label{translation samples}Comparisons of different systems in translating the same example sentence, which from CH-EN test sets. (``source'' indicates the source sentence; ``reference'' indicates the human translation; the translation results are indicated by their system names, including our best ``WP$_\text{ED}$" systems. The underline words in the reference are missed in the baseNMT output; the bold font indicates improvements over the baseNMT system; and the italic font indicates remaining translation errors.) }
\end{table*}

Although the comparison may not be fair enough due to the language pair and training conditions, the above relative improvements (e.g. WP$_{\text{ED}}$-V v.s. baseNMT) is much higher than previous
research of manipulating the vocabularies~\cite{Jean2015On,Haitao2016vocabulary,LHostisGA2016Vocabulary}. This is because our mechanism is not only about reducing the vocabulary itself for each sentence or batch, it also brings improvement to the overall translation model. Please note that unlike these research, we keep the target vocabulary to be 30k in all our experiments, because we are not focusing on increasing the vocabulary size in this paper. It will be interesting to combine our mechanism with larger vocabulary to further enhance the translation performance. Again, our mechanism requires no extra annotation, dictionary, alignment or separate discriminative predictor, etc.

\subsection{Translation Analysis}
We also analyze real-case translations to see the difference between different systems (Table~\ref{translation samples}). 

It is easy to see that the baseNMT system misses the translations of several important words, such as ``advertising", ``1.5", which are marked with underline in the reference. It also wrongly translates the company name ``time warner inc." as the redundant information ``internet company"; ``america online" as ``us line". 

The results of dropout or ensemble show improvement compared to the baseNMT. But they still make mistakes about the translation of ``online" and the company name ``time warner inc.". 

With WP$_\text{ED}$, most of these errors no longer exist, because we force the encoder and decoder to carry the  exact information during translation.

\section{Conclusions}
\label{conclusions}
The encoder-decoder architecture provides a general paradigm for learning machine translation from the source language to the target language. However, due to the large amount of parameters and relatively small training data set, the end-to-end learning of an NMT model may not be able to learn the best solution. We argue that at least part of the problem is caused by the long error back-propagation pipeline of the recurrent structures in multiple time steps, which provides no direct control of the information carried by the hidden states in both the encoder and decoder.

Instead of looking for other annotated data, we notice that the words in the target language sentence could be viewed as a natural annotation. We propose to use the word prediction mechanism to enhance the initial state generated by the encoder and extend the mechanism to control the hidden states of decoder as well. Experiments show promising results on the Chinese-English and German-English translation tasks. As a by-product, the word predictor could be used to improve the efficiency of decoding, which may be crucial for large scale applications.

Our attempts demonstrate that the learning of the large scale neural network systems is still not good enough. In the future, it might be helpful to analyze the benefits of jointly learning other related tasks together with machine translation, to provide further control of the learning process. It is interesting to demonstrate the effectiveness of the proposed mechanism on other sequence to sequence learning tasks as well.

\section*{Acknowledgments}
We would like to thank the anonymous reviewers for their insightful comments. Shujian Huang is the corresponding author. This work is supported by the National Science Foundation of China (No. 61672277, 61300158, 61472183).

\bibliographystyle{emnlp_natbib}

\end{document}